%% file: Camouflageator.tex
\documentclass{article} 
\usepackage{iclr2024_conference,times}

\input{math_commands.tex}

\usepackage[pagebackref=true,breaklinks=true,letterpaper=true,colorlinks,bookmarks=false]{hyperref}
\usepackage{url}
\usepackage{epsfig}
\usepackage{graphicx}
\usepackage{amsmath}
\usepackage{amssymb}
\usepackage{booktabs}
\usepackage{algorithm}
\usepackage{algpseudocode}
\usepackage[table,xcdraw,dvipsnames]{xcolor}
\usepackage{multirow}
\usepackage{subcaption}
\usepackage{enumitem}
\usepackage{balance}
\usepackage{wrapfig}

\makeatletter
\newcommand*\bigcdot{\mathpalette\bigcdot@{.5}}
\newcommand*\bigcdot@[2]{\mathbin{\vcenter{\hbox{\scalebox{#2}{$\m@th#1\bullet$}}}}}
\makeatother

\definecolor{c2}{HTML}{FBD9BD}
\definecolor{c3}{HTML}{fe793d}
\definecolor{c4}{HTML}{eedeb0}
\definecolor{rouse}{rgb}{0.981,0.961,0.941}
\newcommand{\ourCOD}{{Camouflageator }}
\newcommand{\ourdetector}{{ICEG }}

\usepackage[capitalize]{cleveref}
\crefname{section}{Sec.}{Secs.}
\Crefname{section}{Section}{Sections}
\Crefname{table}{Table}{Tables}

\title{Strategic Preys Make Acute Predators: \\ Enhancing Camouflaged Object Detectors \\ by Generating Camouflaged Objects}

\author{
Chunming He$^1$\,,
Kai Li$^2$\thanks{ Corresponding Author, $\dagger$ The work was mainly done when Yulun Zhang was at ETH Z\"{u}rich.}\,~,
Yachao Zhang$^1$\,,
Yulun Zhang$^3$\,, \\
\ \textbf{Chenyu You}$^4$\,\textbf{,} 
\textbf{Zhenhua Guo}$^5$\,\textbf{,}
\textbf{Xiu Li}$^{1}$\protect\footnotemark[1]\,\textbf{,}
\textbf{Martin Danelljan}$^6$\,\textbf{,} 
\textbf{Fisher Yu}$^6$ \\
$^1$Shenzhen International Graduate School, Tsinghua University, \\
$^2$NEC Laboratories America, 
$^3$Shanghai Jiao Tong University, 
$^4$Yale University, \\
$^5$Tianyi Traffic Technology,
$^6$ETH Z\"{u}rich, \\}

\iclrfinalcopy 
\begin{document}

\maketitle

\begin{abstract}
Camouflaged object detection (COD) is the challenging task of identifying camouflaged objects visually blended into surroundings. Albeit achieving remarkable success, existing COD detectors still struggle to obtain precise results in some challenging cases. To handle this problem, we draw inspiration from the prey-vs-predator game that leads preys to develop better camouflage and predators to acquire more acute vision systems and develop algorithms from both the prey side and the predator side. On the prey side, we propose an adversarial training framework, Camouflageator, which introduces an auxiliary generator to generate more camouflaged objects that are harder for a COD method to detect. Camouflageator trains the generator and detector in an adversarial way such that the enhanced auxiliary generator helps produce a stronger detector. On the predator side, we introduce a novel COD method, called Internal Coherence and Edge Guidance (ICEG), which introduces a camouflaged feature coherence module to excavate the internal coherence of camouflaged objects, striving to obtain more complete segmentation results. Additionally, ICEG proposes a novel edge-guided separated calibration module to remove false predictions to avoid obtaining ambiguous boundaries. Extensive experiments show that ICEG outperforms existing COD detectors and Camouflageator is flexible to improve various COD detectors, including ICEG, which brings state-of-the-art COD performance. The code will be available at \url{https://github.com/ChunmingHe/Camouflageator}.
\end{abstract}

\setlength{\abovedisplayskip}{2pt}
\setlength{\belowdisplayskip}{2pt}


\section{Introduction}

The never-ending prey-vs-predator game drives preys to develop various escaping strategies. One of the most effective and ubiquitous strategies is camouflage. Preys use camouflage to blend into the surrounding environment, striving to escape hunting from predators. For survival, predators, on the other hand, must develop acute vision systems to decipher camouflage tricks. 
Camouflaged object detection (COD) is the task that aims to mimic predators' vision systems and localize foreground objects that have subtle differences from the background. 
The intrinsic similarity between camouflaged objects and the backgrounds renders COD a more challenging task than traditional object detection~\citep{liu2020deep}, and has attracted increasing research attention for its potential applications in
medical image analysis~\citep{tang2023source} and species discovery~\citep{he2023degradation}.

Traditional COD solutions~\citep{hou2011detection,pan2011study} mainly rely on manually designed strategies with fixed extractors and thus are constrained by limited discriminability. 
Benefiting from the powerful feature extraction capacity of convolutional neural network, a series of deep learning-based methods have been proposed and have achieved remarkable success on the COD task~\citep{He2023Camouflaged,he2023weaklysupervised,zhaiexploring}.  However, when facing some extreme camouflage scenarios, those methods still struggle to excavate sufficient discriminative cues crucial to \textit{precisely} localize objects of interest. \begin{figure}[htbp]
	\begin{center}	    
	\setlength{\abovecaptionskip}{0.1cm} 
	\setlength{\belowcaptionskip}{-0.2cm}
	\includegraphics[width=0.7\linewidth]{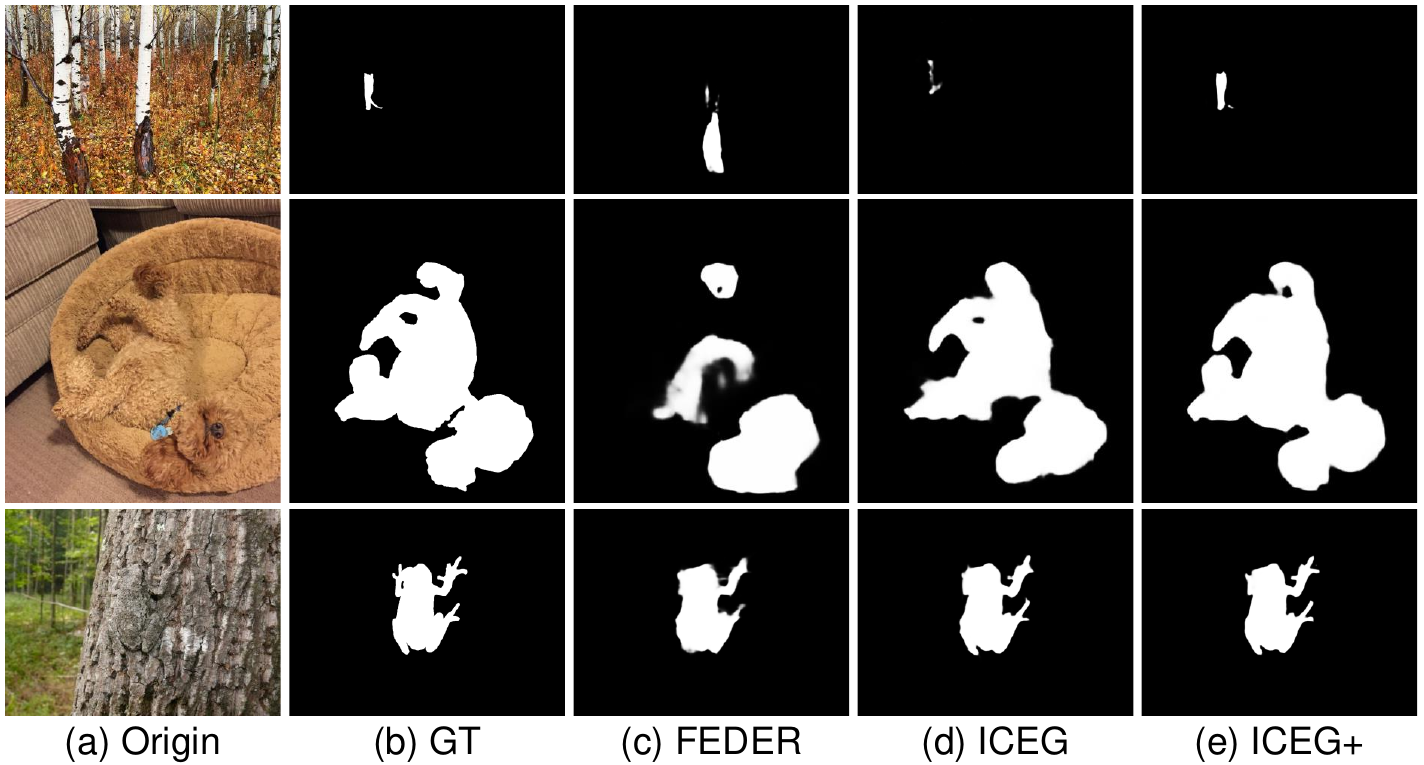} %
 \vspace{0.3mm}
	\caption{ Results of FEDER~\citep{He2023Camouflaged}, ICEG, and ICEG+. ICEG+ indicates to optimize ICEG under the Camouflageator framework. Both ICEG and ICEG+ generate more complete results with clearer edges. ICEG+ also exhibits better localization capacity.} 
 \end{center}
 \label{fig:challenges}
 \vspace{-4mm}
\end{figure}For example, as shown in the top row of Fig.~{\color{red}1}, the state-of-the-art COD method, FEDER \citep{He2023Camouflaged}, cannot even roughly localize the object and thus produce a completely wrong result. Sometimes, even though a rough position can be obtained, FEDER still fails to precisely segment the objects, as shown in the two remaining rows of Fig.~{\color{red}1}. While FEDER manages to find the rough regions for the objects, the results are either incomplete (middle row: some key parts of the dog are missing) or ambiguous 
(bottom row: the boundaries of the frog are not segmented out). 



{This paper aims to address these limitations. We are inspired by the prey-vs-predator game, where preys develop more deceptive camouflage skills to escape predators, which, in turn, pushes the predators to develop more acute vision systems to discern the camouflage tricks. This game leads to ever-strategic preys and ever-acute predators. With this inspiration, we propose to address COD by developing algorithms on both the prey side that generates more deceptive camouflage objects and the predator side that produces complete and precise detection results.  }

{On the prey side, we propose a novel adversarial training framework, Camouflageator, which generates more camouflaged objects that make it even harder for existing detectors to detect and thus enhance the generalizability of the detectors. Specifically, as shown in~\cref{fig:CamouflageatorFramework}, Camouflageator comprises an auxiliary generator and a detector, which could be any existing detector. We adopt an alternative two-phase training mechanism to train the generator and the detector. In Phase \uppercase\expandafter{\romannumeral1}, we freeze the detector and train the generator to synthesize camouflaged objects aiming to deceive the detector. In Phase \uppercase\expandafter{\romannumeral2}, we freeze the generator and train the detector to accurately segment the synthesized camouflaged objects. By iteratively alternating Phases \uppercase\expandafter{\romannumeral1} and \uppercase\expandafter{\romannumeral2}, the generator and detector both evolve, helping to obtain better COD results. }


{On the predator side, we present a novel COD detector, termed Internal Coherence and Edge Guidance (ICEG), which particularly aims to address the issues of incomplete segmentation and ambiguous boundaries of existing COD detectors. For incomplete segmentation, we introduce a camouflaged feature coherence (CFC) module to excavate the internal coherence of camouflaged objects. We first explore the feature correlations using two feature aggregation components, \textit{i.e.}, the intra-layer feature aggregation and the contextual feature aggregation. Then, we propose a camouflaged consistency loss to constrain the internal consistency of camouflaged objects. To eliminate ambiguous boundaries, we propose an edge-guided separated calibration (ESC) module. ESC separates foreground and background features using attentive masks to decrease uncertainty boundaries and remove false predictions. Besides, ESC leverages edge features to adaptively guide segmentation and reinforce the feature-level edge information to achieve the sharp edge for segmentation results. We integrate the Camouflageator framework with ICEG to get ICEG+, which can exhibit better localization capacity (see Fig. ~{\color{red}1}).
Our contributions are summarized as follows:}
\begin{itemize}
    \setlist{nolistsep}
	\vspace{-8pt}
	\item[$\bullet$] We design an adversarial training framework, Camouflageator, for the COD task. Camouflageator employs an auxiliary generator that generates more camouflaged objects that are harder for COD detectors to detect and hence enhances the generalizability of those detectors. Camouflageator is flexible and can be integrated with various existing COD detectors. 
	\vspace{-12pt}
	\item[$\bullet$] We propose a new COD detector, ICEG, to address the issues of incomplete segmentation and ambiguous boundaries that existing detectors face. ICEG introduces a novel CFC module to excavate the internal coherence of camouflaged objects to obtain complete segmentation results, and an ESC module to leverage edge information to get precise boundaries. 
	\vspace{-12pt}
	\item[$\bullet$] Experiments on four datasets verify that Camouflageator can promote the performance of various existing COD detectors, ICEG significantly outperforms existing COD detectors, and integrating Camouflageator with ICEG reaches even better results. 
\end{itemize}

\section{Related work}
\subsection{Camouflaged object detection}
Traditional methods rely on hand-crafted operators with limited feature discriminability~\citep{he2019image}, failing to handle complex scenarios. {A Bayesian-based method~\citep{zhang2016bayesian} was proposed to separate the foreground and background regions through camouflage modeling.} Learning-based approaches have become mainstream in COD with three main categories: \textit{(\romannumeral1) Multi-stage framework:} SegMaR~\citep{jia2022segment} was the first plug-and-play method to integrate segment, magnify, and reiterate under a multi-stage framework. However, SegMaR has limitations in flexibility due to not being end-to-end trainable. \textit{(\romannumeral2) Multi-scale feature aggregation:} PreyNet~\citep{zhang2022preynet} proposed a bidirectional bridging interaction module to aggregate cross-layer features with attentive guidance. {UGTR~\citep{yang2021uncertainty} proposed a probabilistic representational model combined with transformers to explicitly address uncertainties. DTAF~\citep{ren2021deep} developed multiple texture-aware refinement modules to learn the texture-aware features.} Similarly, FGANet~\citep{zhaiexploring} designed a collaborative local information interaction module to aggregate structure context features. \textit{(\romannumeral3) Joint training strategy:} {MGL~\citep{zhai2021mutual} designed the mutual graph reasoning to model the correlations between the segmentation map and the edge map.} BGNet~\citep{sun2022boundary} presented a joint framework for COD to detect the camouflaged candidate and its edge using a cooperative strategy. Analogously, FEDER~\citep{He2023Camouflaged} jointly trained the edge reconstruction task with the COD task and guided the segmentation with the predicted edge. 

We improve existing methods in three aspects: \textit{(\romannumeral1)} 
Camouflageator is the first end-to-end trainable plug-and-play framework for COD, thus ensuring flexibility. \textit{(\romannumeral2)} ICEG is the first COD detector to alleviate incomplete segmentation by excavating the internal coherence of camouflaged objects. \textit{(\romannumeral3)} Unlike existing edge-based detectors~\citep{sun2022boundary,He2023Camouflaged,xiao2023concealed}, ICEG employs edge information to guide segmentation adaptively under the separated attentive framework.

\subsection{Adversarial training}
Adversarial training is a widely-used solution with many applications, including adversarial attack~\citep{zhang2021survey} and generative adversarial network (GAN)~\citep{deng2022pcgan,li2020adversarial}. Recently, several GAN-based methods have been proposed for the COD task. JCOD~\citep{li2021uncertainty} introduced a GAN-based framework to measure the prediction uncertainty. ADENet~\citep{xiang2021exploring} employed GAN to weigh the contribution of depth for COD. Distinct from those GAN-based methods, our Camouflageator enhances the generalizability of existing COD detectors by generating more camouflaged objects that are harder to detect.




\vspace{-1mm}
\section{Methodology}\label{Sec:Method}\vspace{-1mm}
{When preys develop more deceptive camouflaged skills to escape predators, the predators respond by evolving more acute vision systems to discern the camouflage tricks. Drawing inspiration from this prey-vs-predator game, we propose to address COD by developing the Camouflageator and ICEG techniques that mimic preys and predators, respectively, to generate more camouflaged objects and to more accurately detect the camouflaged objects, improving the generalizability of the detector. }

\vspace{-1mm}
\subsection{Camouflageator}\label{Sec:Camouflageator}\vspace{-1mm}

\begin{figure}[tbp]
	\centering
	\setlength{\abovecaptionskip}{0.1cm} 
	\setlength{\belowcaptionskip}{-0.0cm}
	\includegraphics[width=\linewidth]{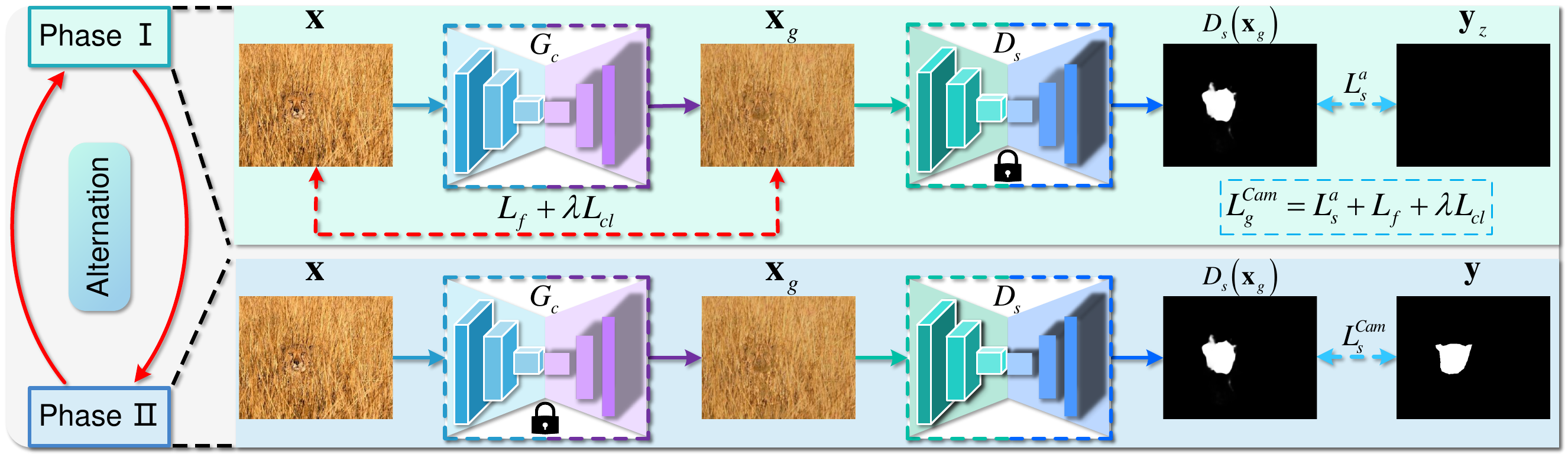} %
	\caption{Architecture of Camouflageator. In Phase \uppercase\expandafter{\romannumeral1}, we fix detector $D_s$ and update generator $G_c$ to synthesize more camouflaged objects to deceive $D_s$. In Phase \uppercase\expandafter{\romannumeral2}, we fix $G_c$ and train the detector $D_s$ to segment the synthesized image.} 
 \vspace{-5mm}
	\label{fig:CamouflageatorFramework}
\end{figure}

Camouflageator is an adversarial training framework that employs an auxiliary generator $G_c$ to synthesize more camouflaged objects that make it even harder for existing detectors $D_s$ to detect and thus enhance the generalizability of the detectors.
We train $G_c$ and $D_s$ alternatively in a two-phase adversarial training scheme. 
\cref{fig:CamouflageatorFramework} shows the framework.

\noindent\textbf{Training the generator.} 
We fix the detector $D_s$ and train the generator $G_c$ to generate more deceptive objects that fail   
the detector. Given a camouflaged image $\mathbf{x}$,
we generate 
\begin{equation}
\mathbf{x}_g=G_c(\mathbf{x}),
\end{equation}
and expect that $\mathbf{x}_g$ is more deceptive to $D_s$ than $\mathbf{x}$. To achieve this, $\mathbf{x}_g$ should be visually consistent (similar in global appearance) with $\mathbf{x}$ but simultaneously have those discriminative features crucial for detection hidden or reduced. 

To encourage visual consistency, we propose to optimize the fidelity loss represented as follows:
\begin{equation}
L_f = \|\left(\mathbf{1}\!-\!\mathbf{y}\right)\!\otimes\! \mathbf{x}_g-\left(\mathbf{1}\!-\!\mathbf{y}\right)\!\otimes\! \mathbf{x}\|^2,
	\label{Eq:fid_loss}
\end{equation}
where $\mathbf{y}$ is the ground truth binary mask and $\otimes$ denotes element-wise multiplication. 
Since $(\mathbf{1}\!-\!\mathbf{y})$ denotes the background mask, this term in essence encourage $\mathbf{x}_g$ to be similar with $\mathbf{x}$ for the background region. We encourage fidelity by preserving only the background rather than the whole image because otherwise, it hinders the generation of camouflaged objects in the foreground.

To hide discriminative features, we optimize the following concealment loss to imitate the bio-camouflage strategies, \textit{i.e.}, internal similarity and edge disruption \citep{price2019background}, as 
\begin{equation}
        L_{cl} = \|\mathbf{y} \otimes\! \mathbf{x}_g-P_o^{I}\|^2 + \!\| \mathbf{y}_e \otimes\! \mathbf{x}_g - P_e^{I} \|^2,	
	\label{Eq:ConcealLoss}
\end{equation}
where $\mathbf{y}_e$ is the weighted edge mask dilated by Gaussian function~\citep{jia2022segment} to capture richer edge information. 
$P_o^{I}$ is the image-level object prototype which is an average of foreground pixels.  
$P_e^{I}$ is the image-level edge prototype which is an average of edge pixels specified by $\mathbf{y}_e$. Note that $\mathbf{y}_e$, $P_o^{I}$, and $P_e^{I}$ are all derived from the provided ground truth $\mathbf{y}$ and help to train the model. This term encourages individual pixels of the foreground region and the edge region of $\mathbf{x}_g$ to be similar to the average values, which has a smooth effect and thus hides discriminative features.

Apart from the above concealment loss, we further employ the detector $D_s$ to reinforce the concealment effect. The idea is that if $\mathbf{x}_g$ is perfectly deceptive, $D_s$ tends to detect nothing as the foreground. To this end, we optimize   
\begin{equation}
\centering
L_{s}^a\!=\!L^w_{BCE}\left(D_s\left(\mathbf{x}_g\right),\mathbf{y}_z\right)\!+\!L^w_{IoU}\left(D_s\left(\mathbf{x}_g\right),\mathbf{y}_z\right),
\label{Eq:AdvLoss}
\end{equation}
where $\mathbf{y}_z\!=\!\mathbf{0}$ is an all-zero mask. $L^w_{BCE}(\bigcdot)$ and $L^w_{IoU}(\bigcdot)$ denote the weighted binary cross-entropy loss~\citep{jadon2020survey} and the weighted intersection-over-union loss~\citep{rahman2016optimizing}. 

By introducing a trade-off parameter $\lambda$, our overall learning objective to train $G_c$ is as follows,
\begin{equation}
	\centering
	L_g^{Cam}=L_s^a + L_f + \lambda L_{cl}.
\label{Eq:GeneLoss}
\end{equation}

\begin{figure*}[htbp]
	\centering
	\setlength{\abovecaptionskip}{0.1cm} 
	\setlength{\belowcaptionskip}{-0.2cm}
	\includegraphics[width=\linewidth]{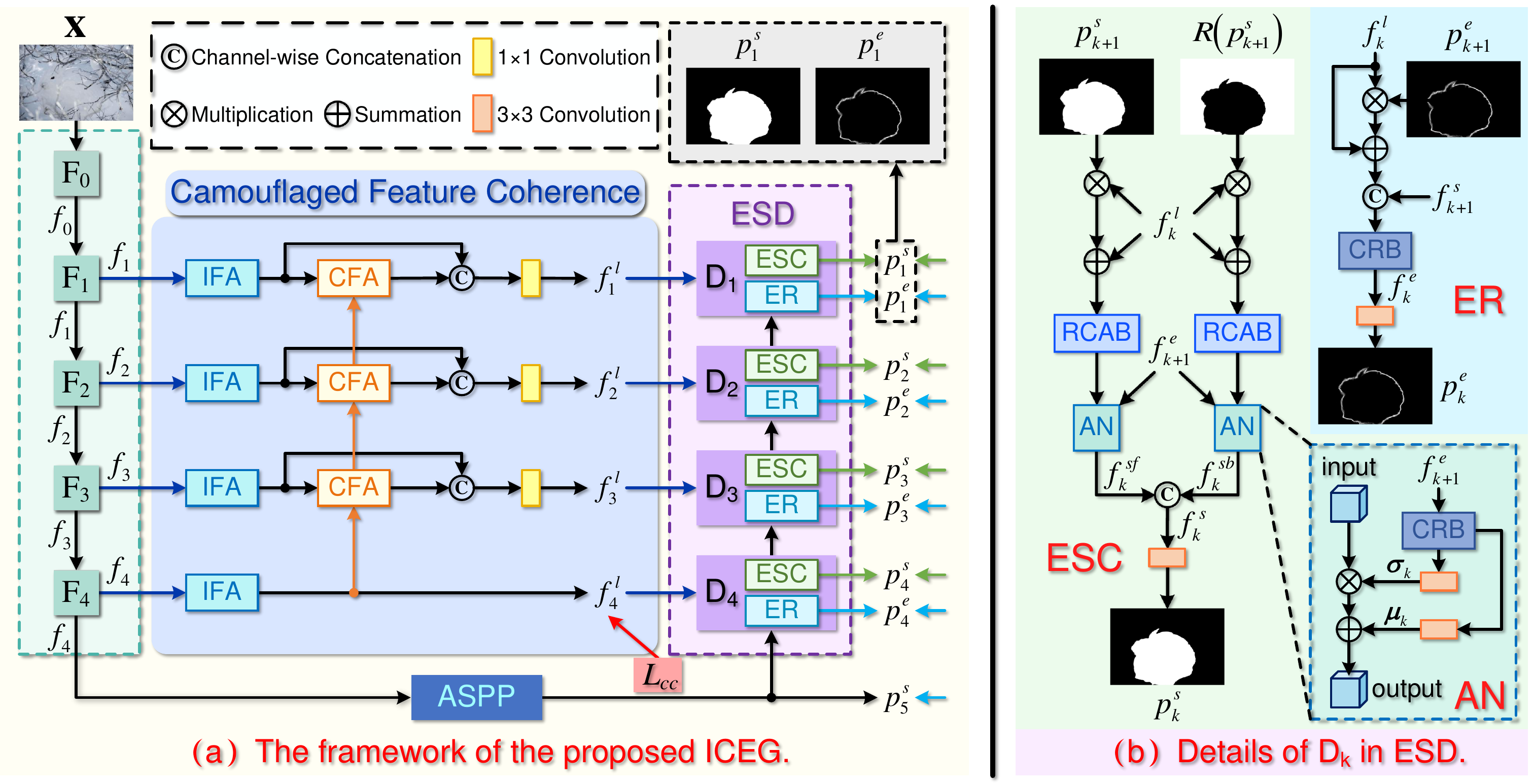}
	\caption{\small Framework of our ICEG. CRB is the Conv-ReLU-BN structure. We omit the Sigmoid operator in (b) for clarity.}\label{fig:ICENetFramework}\vspace{-4mm}
\end{figure*}

\noindent\textbf{Training the detector.}
In Phase \textbf{\uppercase\expandafter{\romannumeral2}}, we fix the generator $G_c$ and train the detector $D_s$ to accurately segment the synthesized camouflaged objects. This is the standard COD task and various existing COD detectors can be employed, for example, the simple one we used above,
\begin{equation}
	\centering
L_{s}^{Cam}\!=\!L^w_{BCE}\left(D_s\left(\mathbf{x}_g\right),\mathbf{y}\right)\!+\!L^w_{IoU}\left(D_s\left(\mathbf{x}_g\right),\mathbf{y}\right).
	\label{Eq:AdvLoss1}
\end{equation}


\subsection{ICEG}\label{Sec:detector}
{We further propose ICEG to alleviate incomplete segmentation and eliminate ambiguous boundaries.}
Given $\mathbf{x}$ of size $W\times H$, we start by using a basic encoder $F$ to extract a set of deep features $\{f_k\}_{k=0}^4$ with the resolution of $\frac{W}{2^{k+1}}\times \frac{H}{2^{k+1}}$ and employ ResNet50~\citep{he2016deep} as the default architecture.
As shown in~\cref{fig:ICENetFramework}, we then feed these features, i.e., $\{f_k\}_{k=1}^4$, to the camouflaged feature coherence (CFC) module and the edge-guided segmentation decoder (ESD) for further processing. Moreover, 
the last feature map $f_4$, which has rich semantic cues, 
is fed into an atrous spatial pyramid pooling (ASPP) module $A_s$~\citep{yang2018denseaspp} and a $3\times3$ convolution $conv3$ to generate a coarse result $p^s_5$: $p^s_5=conv3(A_s(f_4))$, where $p^s_5$ shares the same spatial resolution with $f_4$.

\vspace{-1mm}
\subsubsection{Camouflaged feature coherence module}\label{Sec:FAmodules} 
\vspace{-1mm}
{To alleviate incomplete segmentation, we propose the camouflaged feature coherence (CFC) module }
\begin{wrapfigure}[7]{r}{0.45\textwidth}
\vspace{-5mm}
\centering
	\setlength{\abovecaptionskip}{0cm} 
	\setlength{\belowcaptionskip}{0cm}
	\includegraphics[width=\linewidth]{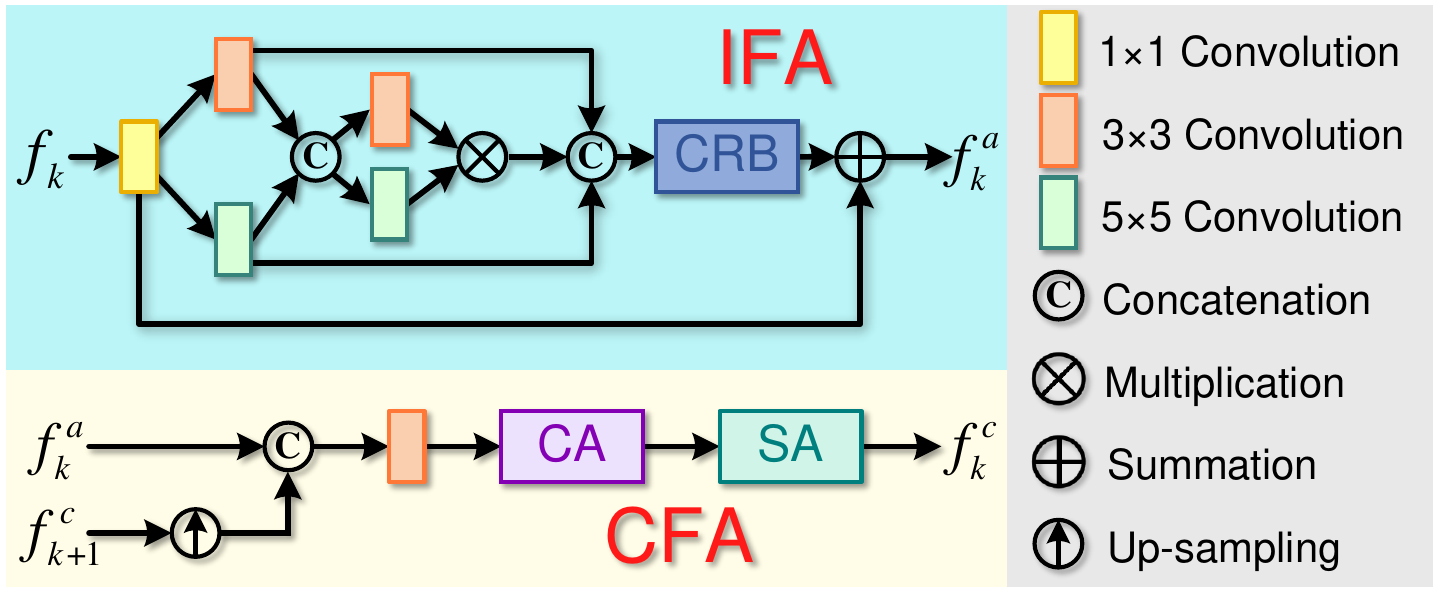} 
	\caption{\small Details of IFA and CFA.}
 \vspace{-4mm}
	\label{fig:IFA+CFA}
\end{wrapfigure}{to excavate the internal coherence of camouflaged objects. }
CFC consists of two feature aggregation components, \textit{i.e.}, the intra-layer feature aggregation (IFA) and the contextual feature aggregation (CFA), to explore feature correlations. Besides, CFC introduces a camouflaged consistency loss to constrain the internal consistency of camouflaged objects.


\noindent \textbf{Intra-layer feature aggregation.} In~\cref{fig:IFA+CFA}, IFA seeks the feature correlations by integrating the multi-scale features with different reception fields in a single layer, assuring that the aggregated features can capture scale-invariant information.
Given $f_k$, a $1\times 1$ convolution $conv1$ is first applied for channel reduction, followed by two parallel convolutions with different kernel sizes. This process produces the features $f_k^3$ and $f_k^5$ with varying receptive fields:
\begin{equation}
	f_k^3=conv3(conv1(f_k)), f_k^5=conv5(conv1(f_k)),
\end{equation}
where $conv5$ is $5\times 5$ convolution. Then we combine $f_k^3$ and $f_k^5$, process them with two parallel convolutions, and multiply the outputs to excavate the scale-invariant information:
\begin{equation}
	\hspace{-3mm}
	f_k^{35}\!=conv3\!\left(conca\!\left(f_k^3,f_k^5\right)\!\right)\!\otimes  conv5\left(conca\!\left(f_k^3,f_k^5\right)\!\right)\!,
\end{equation}
where $conca(\bigcdot)$ denote concatenation. We then integrate the three features and process them with a CRB block $CRB(\bigcdot)$, \textit{i.e.}, $3\times 3$ convolution, ReLU, and batch normalization. By summing with the channel-wise down-sampled feature, the aggregated features $\{f_k^a\}_{k=1}^4$ are formulated as follows:
\begin{equation}
	f_k^a=conv1(f_k)+ CRB\left(conca\left(f_k^3,f_k^5,f_k^{35}\right)\right).
\end{equation}
\noindent \textbf{Contextual feature aggregation.} CFA explores the inter-layer feature correlations by selectively interacting cross-level information with channel attention and spatial attention~\citep{woo2018cbam}, which ensures the retention of significant coherence. The aggregated feature $\{f_k^c\}_{k=1}^3$ is
\begin{equation}
	f_k^c=SA\left(CA\left(conv3\left(conca\left(up\left(f_{k+1}^c\right),f_k^a\right)\right)\right)\right),
\end{equation}
where $up(\bigcdot)$ is up-sampling operation. $CA\left(\bigcdot\right)$ and $SA\left(\bigcdot\right)$ are channel attention and spatial attention. $f_4^c=f_4^a$. Given $\{f_k^c\}_{k=1}^3$, the integrated features $\{f_k^l\}_{k=1}^3$ conveyed to the decoder are
\begin{equation}
	f_k^l=conv1\left(concate\left(f_k^a,f_k^c\right)\right).
\end{equation}
We employ $conv1$ for channel integration and $f_4^l=f_4^a$.


\noindent \textbf{Camouflaged consistency loss.} To enforce the internal consistency of the camouflaged object, we propose a camouflaged consistency loss to enable more compact internal features. To achieve this, one intuitive idea is to decrease the variance of the camouflaged internal features. However, such a constraint can lead to feature collapse, \textit{i.e.}, all extracted features are too clustered to be separated, thus diminishing the segmentation capacity. Therefore, apart from the above constraint, we propose an extra requirement 
to keep the internal and external features as far away as possible. We apply the feature-level consistency loss to the deepest feature $f_4^l$ for its abundant semantic information:
\begin{equation}
	L_{cc}\!=\! \|\mathbf{y}_d \otimes\! f_4^l-P_o^{f}\|^2\!-\!\|\mathbf{y}_d \otimes\! f_4^l-P_b^{f}\|^2,
	\label{Eq:CCLoss}
\end{equation}
where $\mathbf{y}_d$ is the down-sampled ground truth mask. $P_o^{f}$ and $P_b^{f}$ denote the feature-level prototypes of the camouflaged object and the background, respectively.

\noindent \textbf{Discussions.} Apart from focusing on feature correlations as in existing detectors~\citep{zhang2022preynet,He2023Camouflaged}, we design a novel camouflaged consistency loss to enhance the internal consistency of camouflaged objects, facilitating complete segmentation.

\vspace{-2mm}
\subsubsection{Edge-guided segmentation decoder}\label{Sec:Decoder}
As depicted in~\cref{fig:ICENetFramework}, edge-guided segmentation decoder (ESD) $\{D_k\}_{k=1}^4$
comprises an edge reconstruction (ER) module and an edge-guided separated calibration (ESC) module to generate the edge predictions $\{p_k^e\}_{k=1}^4$ and the segmentation results $\{p_k^s\}_{k=1}^5$, respectively.

\noindent \textbf{Edge reconstruction module.} 
We introduce an ER module to reconstruct the object boundary. Assisted by the edge map $p_{k+1}^e$ and the segmentation feature $f_{k+1}^s$ from the former decoder, the edge feature $f_k^e$ is presented as follows: 
\begin{equation}
	\begin{aligned}
	f_k^e& = CRB(conca(f_k^l\otimes p_{k+1}^e+f_k^l,f_{k+1}^s)).
	\end{aligned}
\label{Eq:ER}
\end{equation}
where $f_5^s=A_s(f_4)$ and $p_k^e=conv3(f_k^e)$. $f_5^e$ and $p_5^e$ are set as zero for initialization. 
We repeat $p_{k+1}^e$ as a 64-dimension tensor to ensure channel consistency with $f_k^l$ in~\cref{Eq:ER}.

\noindent \textbf{Edge-guided separated calibration module.} 
{Ambiguous boundary, a common problem in COD, manifests as two phenomena: (1) a high degree of uncertainty in the fringes, and (2) the unclear edge of the segmented object.} We have observed that the high degree of uncertainty is mainly due to the intrinsic similarity between the camouflaged object and the background.
To address this issue, we separate the features from the foreground and the background by introducing the corresponding attentive masks,
and design a two-branch network to process the attentive features. This approach helps decrease uncertainty fringes and remove false predictions, including false-positive and false-negative errors. Given the prediction map $p_{k+1}^s$, the network is defined as follows:
\begin{equation}
	\begin{aligned}
		f_k^{s}=conca(f_k^{sf},f_k^{sb}), p_k^{s}=conv3(f_k^{s}),   \\
	\end{aligned}
	\label{Eq:ESC1}
\end{equation}
where $f_k^{sf}$ and $f_k^{sb}$ are the foreground and the background attentive features, which are formulated
\begin{subequations}
	\begin{align}
		&f_k^{sf}=RCAB\left(f_k^l\otimes S\left(p_{k+1}^s\right)+f_k^l\right),  \\
		&f_k^{sb}=RCAB\left(f_k^l\otimes S\left(R\left(p_{k+1}^s\right)\right)+f_k^l\right),
	\end{align}
	\label{Eq:ESC2}%
\end{subequations}
where $S(\bigcdot)$ and $R(\bigcdot)$ are Sigmoid and reverse operators, \textit{i.e.}, element-wise subtraction with 1. $RCAB(\bigcdot)$ is the residual channel attention block~\citep{zhang2018image}, which is used to emphasize those informative channels and high-frequency information. 

The second phenomenon, unclear edge, is due to the extracted features giving insufficient importance to edge information. In this case, we explicitly incorporate edge features to guide the segmentation process and promote edge prominence.
Instead of simply superimposing, we design an adaptive normalization (AN) strategy with edge features to guide the segmentation in a variational manner, which reinforces the feature-level edge information and thus ensures the sharp edge of the segmented object. Given the edge feature $f_{k+1}^e$, the attentive features can be acquired by:
\begin{subequations}
	\hspace{-5mm}
	\begin{align}
		&f_k^{sf}\!=\!\boldsymbol{\sigma}_k^f\!\otimes\!(RCAB(f_k^l\!\otimes\! S(p_{k+1}^s)\!+\!f_k^l))\!+\!\boldsymbol{\mu}_k^f, \label{Eq:ESC3-1} \\
		&f_k^{sb}\!=\!\boldsymbol{\sigma}_k^b\!\otimes\!(RCAB(f_k^l\!\otimes\! S(R(p_{k+1}^s))\!+\!f_k^l))\!+\!\boldsymbol{\mu}_k^b, \label{Eq:ESC3-2}
	\end{align}
	\label{Eq:ESC3}%
\end{subequations}
where $\{\boldsymbol{\sigma}_k^f,\boldsymbol{\mu}_k^f\}$ and $\{\boldsymbol{\sigma}_k^b,\boldsymbol{\mu}_k^b\}$ are 
variational parameters. In AN, $\left\{\boldsymbol{\sigma}_k,\boldsymbol{\mu}_k\right\}$ can be calculated by:
\begin{equation}
	\hspace{-3mm}
	\begin{aligned}
		\boldsymbol{\sigma}_k\!=\!conv3_\sigma(CRB_\sigma(f_{k+1}^e)\!), \boldsymbol{\mu}_k\!=\!conv3_\mu(CRB_\mu(f_{k+1}^e)\!).  \\
	\end{aligned}
	\label{Eq:ESC4}
\end{equation}

\noindent \textbf{Discussions.} Unlike existing edge-guided methods~\citep{sun2022boundary,He2023Camouflaged} that focus only on edge guidance, we combine edge guidance with foreground/background splitting using attentive masks. This integration enables us to decrease uncertainty fringes and remove false predictions along edges, thus achieving the sharp edge for segmentation results.

\subsubsection{Loss functions of ICEG} \label{Sec:loss}
Apart from the camouflaged consistency loss $L_{cc}$, our \ourdetector is also constraint with the segmentation loss $L_s$ and the edge loss $L_e$ to supervise the segmentation results $\{p_k^s\}_{k=1}^5$ and the reconstructed edge results $\{p_k^e\}_{k=1}^4$. Following~\citep{fan2021concealed}, we define $L_s$ as
\begin{equation}
	\hspace{-3mm}
	\setlength{\abovedisplayskip}{0pt}
	\setlength{\belowdisplayskip}{0pt}
	L_{s}\!=\!\sum_{k=1}^{5}\frac{1}{2^{k-1}}\left(L^w_{BCE}\left(p_k^s,\mathbf{y}\right)\!+\!L^w_{IoU}\left(p_k^s,\mathbf{y}\right)\right).
\end{equation}
For edge supervision, we employ dice loss $L_{dice}(\bigcdot)$~\citep{milletari2016v} to overcome the extreme imbalance in edge maps:
\begin{equation}
	\setlength{\abovedisplayskip}{0pt}
	\setlength{\belowdisplayskip}{0pt}
	L_e=\sum_{k=1}^{4}\frac{1}{2^{k-1}}L_{dice}\left(p_k^e, \mathbf{y}_e\right).
\end{equation}
Therefore, with the assistance of a trade-off parameter $\beta$, the total loss is presented as follows:
\begin{equation}
	L_t = L_s+ L_e+ \beta L_{cc}.
	\label{Eq:ICENetLoss}
\end{equation}
\subsubsection{ICEG+} \label{Sec:Adversarial}
To promote the adoption of our Camouflageator, we provide a use case and utilize ICEG+ to denote the algorithm that integrates our \ourCOD framework with ICEG. The integration is straightforward; we only need to replace the detector supervision from~\cref{Eq:AdvLoss1} with~\cref{Eq:ICENetLoss}. In addition, we pre-train ICEG with $L_t$ (\cref{Eq:ICENetLoss}) to ensure the training stability. See~\cref{Sec:ExperimentSet} for more details.
\section{Experiments}
\subsection{Experimental setup}\label{Sec:ExperimentSet}
\noindent \textbf{Implementation details.} 
All experiments are implemented on PyTroch on two RTX3090 GPUs. For Camouflageator, the generator adopts ResUNet as its backbone. As for ICEG, a pre-trained ResNet50~\citep{he2016deep} on ImageNet~\citep{krizhevsky2017imagenet} is employed as the default encoder. We also report the COD results with other encoders, including Res2Net50~\citep{gao2019res2net} and Swin Transformer~\citep{liu2021swin}.
Following~\citep{fan2020camouflaged}, we resize the input image as $352\times352$ and pre-train ICEG by Adam with momentum terms $\left(0.9,0.999\right)$ for 100 epochs. The batch size is set as 36 and the learning rate is initialized as 0.0001, decreased by 0.1 every 50 epochs. Then we use the same batch size to further optimize ICEG under the Camouflageator framework for 30 epochs and get ICEG+, where the optimizer is Adam with parameters $\left(0.5,0.99\right)$ and the initial learning rate is 0.0001, dividing by 10 every 15 epochs. 
$\lambda$ and $\beta$ are set as 0.1.
\begin{table*}[htbp!]
	\setlength{\abovecaptionskip}{0.1cm} 
	\setlength{\belowcaptionskip}{-0.2cm}
	\centering
	\resizebox{\columnwidth}{!}{
		\setlength{\tabcolsep}{1.4mm}
		\begin{tabular}{lccccccccccccccccc} 
			\toprule[1.5pt]
			\multicolumn{1}{c|}{}                                        & \multicolumn{1}{c|}{}                           & \multicolumn{4}{c|}{\textit{CHAMELEON} }                                                                                                                                         & \multicolumn{4}{c|}{\textit{CAMO} }                                                                                                                                             & \multicolumn{4}{c|}{\textit{COD10K} }                                                                                                                                          & \multicolumn{4}{c}{\textit{NC4K} }                                                                                                                        \\ \cline{3-18} 
			\multicolumn{1}{l|}{\multirow{-2}{*}{Methods}} & \multicolumn{1}{c|}{\multirow{-2}{*}{Backbones}} & {\cellcolor{gray!40}$M$~$\downarrow$}                                  & {\cellcolor{gray!40}$F_\beta$~$\uparrow$}                               & {\cellcolor{gray!40}$E_\phi$~$\uparrow$}                               & \multicolumn{1}{c|}{\cellcolor{gray!40}$S_\alpha$~$\uparrow$}                                   & {\cellcolor{gray!40}$M$~$\downarrow$}                                  & {\cellcolor{gray!40}$F_\beta$~$\uparrow$}                               & {\cellcolor{gray!40}$E_\phi$~$\uparrow$}                               & \multicolumn{1}{c|}{\cellcolor{gray!40}$S_\alpha$~$\uparrow$}                                   & {\cellcolor{gray!40}$M$~$\downarrow$}                                  & {\cellcolor{gray!40}$F_\beta$~$\uparrow$}                               & {\cellcolor{gray!40}$E_\phi$~$\uparrow$}                               & \multicolumn{1}{c|}{\cellcolor{gray!40}$S_\alpha$~$\uparrow$}                                   & {\cellcolor{gray!40}$M$~$\downarrow$}                                  & {\cellcolor{gray!40}$F_\beta$~$\uparrow$}                               & {\cellcolor{gray!40}$E_\phi$~$\uparrow$}                               & \multicolumn{1}{c}{\cellcolor{gray!40}$S_\alpha$~$\uparrow$}                                   \\ \midrule 
			\multicolumn{18}{c}{Common Setting: Single Input Scale and Single Stage}                                   \\ \midrule
			\multicolumn{1}{l|}{SegMaR-1~\citep{jia2022segment}}                 & \multicolumn{1}{c|}{ResNet50}                   & 0.028 & 0.828                                 &  0.944 & \multicolumn{1}{c|}{{0.892}} & 0.072 & {0.772} & 0.861 & \multicolumn{1}{c|}{{\color[HTML]{00B0F0} \textbf{0.805}}} & 0.035                                 & 0.699                                 & 0.890 & \multicolumn{1}{c|}{0.813}                                 & 0.052                                 & 0.767                                 & 0.885                                 & 0.835                                 \\
			\multicolumn{1}{l|}{PreyNet~\citep{zhang2022preynet}}                     & \multicolumn{1}{c|}{ResNet50}                   & {\color[HTML]{FF0000}\textbf{0.027}}                                 & {{0.844}}                                 & {\color[HTML]{00B0F0}\textbf{0.948}}                                 & \multicolumn{1}{c|}{0.895}                                 & 0.077                                 & 0.763                                 & 0.854                                 & \multicolumn{1}{c|}{0.790}                                 & 0.034                                 & {\color[HTML]{00B0F0}\textbf{0.715}}                                 & {\color[HTML]{00B0F0}\textbf{0.894}}                                 & \multicolumn{1}{c|}{0.813}                                 & 0.047                                 & 0.798                                 & 0.887                                 & 0.838                                 \\
			\multicolumn{1}{l|}{FGANet~\citep{zhaiexploring}}                     & \multicolumn{1}{c|}{ResNet50}                   & 0.030                                 & 0.838                                 & 0.944                                 & \multicolumn{1}{c|}{\color[HTML]{00B0F0}\textbf{0.896}}                                 & {\color[HTML]{00B0F0}\textbf{0.070}}                                & 0.769                                 & 0.865                                 & \multicolumn{1}{c|}{0.800}                                 & {\color[HTML]{00B0F0}\textbf{0.032}}                                 & 0.708                                 & {\color[HTML]{00B0F0}\textbf{0.894}}                                 & \multicolumn{1}{c|}{0.803}                                 & 0.047                                 & 0.800                                 & 0.891                                 & 0.837                                 \\
            \multicolumn{1}{l|}{FEDER~\citep{He2023Camouflaged}} & \multicolumn{1}{c|}{ResNet50}  & 0.028 & {\color[HTML]{00B0F0}\textbf{0.850}} & 0.944 & \multicolumn{1}{c|}{0.892} & {\color[HTML]{00B0F0}\textbf{0.070}} & {\color[HTML]{00B0F0}\textbf{0.775}} & {\color[HTML]{00B0F0}\textbf{0.870}} & \multicolumn{1}{c|}{0.802} & {\color[HTML]{00B0F0}\textbf{0.032}} & {\color[HTML]{00B0F0}\textbf{0.715}} & 0.892 & \multicolumn{1}{c|}{0.810} & {\color[HTML]{00B0F0}\textbf{0.046}} & {\color[HTML]{00B0F0}\textbf{0.808}} & {\color[HTML]{00B0F0}\textbf{0.900}} & {\color[HTML]{00B0F0}\textbf{0.842}} \\
			\multicolumn{1}{l|}{ICEG (Ours)}                                                 & \multicolumn{1}{c|}{ResNet50}                   & {\color[HTML]{FF0000} \textbf{0.027}} & {\color[HTML]{FF0000} \textbf{0.858}} & {\color[HTML]{FF0000} \textbf{0.950}} & \multicolumn{1}{c|}{{\color[HTML]{FF0000} \textbf{0.899}}} & {\color[HTML]{FF0000} \textbf{0.068}} & {\color[HTML]{FF0000} \textbf{0.789}} & {\color[HTML]{FF0000} \textbf{0.879}} & \multicolumn{1}{c|}{{\color[HTML]{FF0000} \textbf{0.810}}} & {\color[HTML]{FF0000} \textbf{0.030}} & {\color[HTML]{FF0000} \textbf{0.747}} & {\color[HTML]{FF0000} \textbf{0.906}} & \multicolumn{1}{c|}{{\color[HTML]{FF0000} \textbf{0.826}}} & {\color[HTML]{FF0000} \textbf{0.044}} & {\color[HTML]{FF0000} \textbf{0.814}} & {\color[HTML]{FF0000} \textbf{0.908}} & {\color[HTML]{FF0000} \textbf{0.849}} \\ \hline
            \rowcolor{c2!20}\multicolumn{1}{l|}{PreyNet+ (Ours)}      & \multicolumn{1}{c|}{ResNet50}  & 0.027   & 0.856     & 0.954     & \multicolumn{1}{c|}{0.901}   & 0.074  & 0.778    & 0.869    & \multicolumn{1}{c|}{0.808}  & 0.031 & 0.744  & 0.908   & \multicolumn{1}{c|}{0.833} & 0.044   & 0.821    & 0.912    & 0.859  \\
            \rowcolor{c2!20}\multicolumn{1}{l|}{FGANet+ (Ours)}         & \multicolumn{1}{c|}{ResNet50}  & 0.029   & 0.847     & 0.948     & \multicolumn{1}{c|}{0.899}   & 0.069  & 0.781    & 0.877    & \multicolumn{1}{c|}{0.814}  & 0.030 & 0.735 & 0.911   & \multicolumn{1}{c|}{0.823} & 0.045   & 0.814    & 0.905    & 0.854  \\
            \rowcolor{c2!20}\multicolumn{1}{l|}{FEDER+ (Ours)}         & \multicolumn{1}{c|}{ResNet50}  & 0.027                & 0.855                & 0.947                & \multicolumn{1}{c|}{0.895}                & 0.068                & 0.793                & 0.883                & \multicolumn{1}{c|}{0.820}                & 0.030                & 0.739                & 0.905                & \multicolumn{1}{c|}{0.831}  & 0.043 &0.820 &0.910 &0.845\\
            \rowcolor{c2!20}\multicolumn{1}{l|}{ICEG+ (Ours)}   & \multicolumn{1}{c|}{ResNet50} & 0.026   & 0.863     & 0.952     & \multicolumn{1}{c|}{0.903}   & 0.066  & 0.805    & 0.891    & \multicolumn{1}{c|}{0.829}  & 0.028 & 0.763 & 0.920   & \multicolumn{1}{c|}{0.843} & 0.041   & 0.835    & 0.922    & 0.869  \\ \midrule
			\multicolumn{1}{l|}{SINet V2~\citep{fan2021concealed}}                  & \multicolumn{1}{c|}{Res2Net50}                  & 0.030                                  & 0.816                                 & 0.942                                 & \multicolumn{1}{c|}{0.888}                                 & 0.070                                  & 0.779                                 & 0.882                                 & \multicolumn{1}{c|}{0.822}                                 & 0.037                                 & 0.682                                 & 0.887                                 & \multicolumn{1}{c|}{0.815}                                 & 0.048                                 & 0.792                                 & 0.903                                 & 0.847                                 \\
			\multicolumn{1}{l|}{BGNet~\citep{sun2022boundary}}                   & \multicolumn{1}{c|}{Res2Net50}                   & 0.029                                 & 0.835                                 & 0.944                                 & \multicolumn{1}{c|}{0.895}                                 & 0.073                                 & 0.744                                 & 0.870                                 & \multicolumn{1}{c|}{0.812}                                 & 0.033                                 & 0.714                                 & 0.901                                 & \multicolumn{1}{c|}{0.831}                                 & 0.044                                 & 0.786                                 & 0.907                                 & 0.851                                 \\
			\multicolumn{1}{l|}{ICEG (Ours)}                       & \multicolumn{1}{c|}{Res2Net50}                  & \textbf{0.025}                        & \textbf{0.869}                             & \textbf{0.958}                             & \multicolumn{1}{c|}{\textbf{0.908}}                        & \textbf{0.066}                        & \textbf{0.808}                             & \textbf{0.903}                             & \multicolumn{1}{c|}{\textbf{0.838}}                        & \textbf{0.028}                        & \textbf{0.752}                             & \textbf{0.914}                             & \multicolumn{1}{c|}{\textbf{0.845}}                        & \textbf{0.042}                        & \textbf{0.828}                             & \textbf{0.917}                             & \textbf{0.867}                        \\ \hline
            \rowcolor{c2!20}\multicolumn{1}{l|}{ICEG+ (Ours)}  & \multicolumn{1}{c|}{Res2Net50}                     & 0.023   & 0.873     & 0.960     & \multicolumn{1}{c|}{0.910}   & 0.064  & 0.826    & 0.912    & \multicolumn{1}{c|}{0.845}  & 0.026 & 0.770                        & 0.925   & \multicolumn{1}{c|}{0.853} & 0.040   & 0.844    & 0.928    & 0.878  \\ \midrule
			\multicolumn{1}{l|}{ICON~\citep{zhuge2022salient}}                   & \multicolumn{1}{c|}{Swin }                  & 0.029                                 & 0.848                                 & 0.940                                 & \multicolumn{1}{c|}{0.898}                                 & 0.058                                 & 0.794                                 & 0.907                                 & \multicolumn{1}{c|}{0.840}                                 & 0.033                                 & 0.720                                 & 0.888                                & \multicolumn{1}{c|}{0.818}                                 & 0.041                                 & 0.817                                 & 0.916                                 & 0.858                                 \\ 
			\multicolumn{1}{l|}{ICEG (Ours)}                 & \multicolumn{1}{c|}{Swin}                  & \textbf{0.023}                                 & \textbf{0.860}                                 & \textbf{0.959}                                 & \multicolumn{1}{c|}{\textbf{0.905}}                                 & \textbf{0.044}                                 & \textbf{0.855}                                 & \textbf{0.926}                                 & \multicolumn{1}{c|}{\textbf{0.867}}                                 & \textbf{0.024}                                 & \textbf{0.782}                                 & \textbf{0.930}                                & \multicolumn{1}{c|}{\textbf{0.857}}                                 & \textbf{0.034}                                 & \textbf{0.855}                                 & \textbf{0.932}                                 & \textbf{0.879}                                 \\ \hline
            \rowcolor{c2!20}\multicolumn{1}{l|}{ICEG+ (Ours)} & \multicolumn{1}{c|}{Swin} & 0.022   & 0.867     & 0.961     & \multicolumn{1}{c|}{0.908}   & 0.042  & 0.861    & 0.931    & \multicolumn{1}{c|}{0.871}  & 0.023 & 0.788                        & 0.934   & \multicolumn{1}{c|}{0.862} & 0.033   & 0.861    & 0.937    & 0.883  \\ \midrule
			\multicolumn{18}{c}{Other Setting: Multiple Input Scales (MIS)} \\ \midrule
			\multicolumn{1}{l|}{ZoomNet~\citep{pang2022zoom}}            & \multicolumn{1}{c|}{ResNet50}                   & 0.024                                 & 0.858                                 & 0.943                                 & \multicolumn{1}{c|}{0.902}                                 & 0.066                                 & 0.792                                 & 0.877                                 & \multicolumn{1}{c|}{0.820}                                 & 0.029                                 & 0.740                                 & 0.888                                 & \multicolumn{1}{c|}{0.838}                        & 0.043                                 & 0.814                                 & 0.896                                 & 0.853                                 \\
			\multicolumn{1}{l|}{ICEG (Ours)}          & \multicolumn{1}{c|}{ResNet50}                   & \textbf{0.023}                        & \textbf{0.864}                        & \textbf{0.957}                        & \multicolumn{1}{c|}{\textbf{0.905}}                        & \textbf{0.063}                        & \textbf{0.802}                        & \textbf{0.889}                        & \multicolumn{1}{c|}{\textbf{0.833}}                        & \textbf{0.028}                        & \textbf{0.751}                        & \textbf{0.913}                        & \multicolumn{1}{c|}{\textbf{0.840}}                                 & \textbf{0.042}                        & \textbf{0.827}                        & \textbf{0.911}                        & \textbf{0.873}                        \\ \midrule
			\multicolumn{18}{c}{Other Setting: Multiple Stages (MS)}  \\ \midrule
			\multicolumn{1}{l|}{SegMaR-4~\citep{jia2022segment}}   & \multicolumn{1}{c|}{ResNet50}                   & 0.025                        & 0.855                                 & 0.955                                 & \multicolumn{1}{c|}{0.906}                                 & 0.071                                 & 0.779                                 & 0.865                                 & \multicolumn{1}{c|}{0.815}                                 & 0.033                                 & 0.737                                 & 0.896                                 & \multicolumn{1}{c|}{0.833}                                 & 0.047                                 & 0.793                                 & 0.892                                 & 0.845                                 \\
			\multicolumn{1}{l|}{ICEG-4 (Ours)}         & \multicolumn{1}{c|}{ResNet50}                   & \textbf{0.024}                        & \textbf{0.870}                        & \textbf{0.961}                        & \multicolumn{1}{c|}{\textbf{0.907}}                        & \textbf{0.067}                        & \textbf{0.802}                        & \textbf{0.884}                        & \multicolumn{1}{c|}{\textbf{0.823}}                        & \textbf{0.028}                        & \textbf{0.755}                        & \textbf{0.920}                        & \multicolumn{1}{c|}{\textbf{0.843}}                        & \textbf{0.043}                        & \textbf{0.824}                        & \textbf{0.915}                        & \textbf{0.860}         \\ \bottomrule[1.5pt]              
	\end{tabular}}
	\caption{\small Quantitative comparisons of ICEG and other 13 SOTAs on four benchmarks. 
    SegMaR-1 and SegMaR-4 are SegMaR at one stage and four stages. ``+'' indicates optimizing the detector under our Camouflageator framework. Swin and PVT denote Swin Transformer and PVT V2. The best results are marked in \textbf{bold}. For ResNet50 backbone in the common setting, the best two results are in {\color[HTML]{FF0000} \textbf{red}} and {\color[HTML]{00B0F0} \textbf{blue}} fonts.} \label{table:Qualitative}
	\vspace{-0.2cm}
\end{table*}
\begin{figure*}[t]
	\centering
	\setlength{\abovecaptionskip}{-0.1cm}
	\begin{center}
		\includegraphics[width=\linewidth]{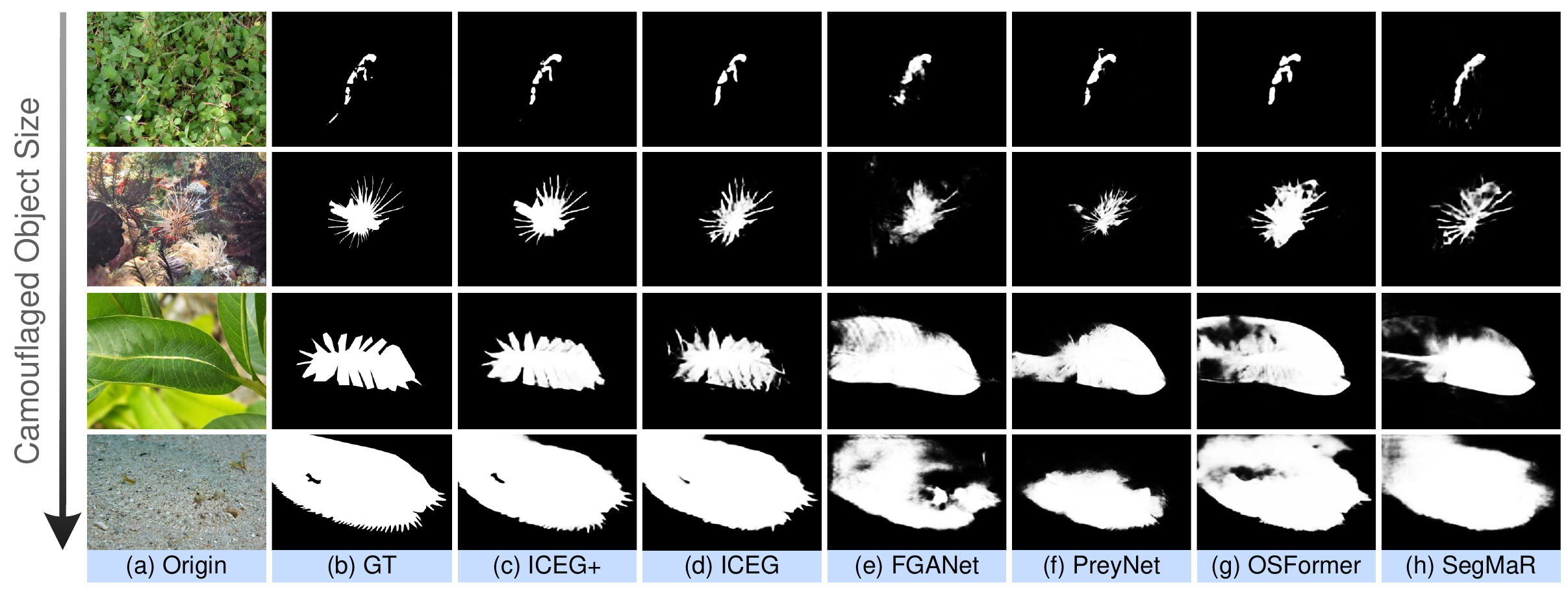}
	\end{center} 
	\caption{\small Qualitative analysis of ICEG and other four cutting-edge methods. ICEG generates more complete results with clearer edges. We also provide the results of ICEG+, which is optimized under Camouflageator. 
	}
	\label{fig:qualitative}
	\vspace{-5mm}
\end{figure*}

\noindent \textbf{Datasets.} We use four COD datasets for evaluation, including \textit{CHAMELEON}~\citep{skurowski2018animal}, \textit{CAMO}~\citep{le2019anabranch},  \textit{COD10K}~\citep{fan2021concealed}, and \textit{NC4K}~\citep{lv2021simultaneously}. \textit{CHAMELEON} comprises 76 camouflaged images. \textit{CAMO} contains 1,250 images with 8 categories.  \textit{COD10K} has 5,066 images with 10 super-classes. \textit{NC4K} is the largest test set with 4,121 images. Following the common setting~\citep{fan2020camouflaged,fan2021concealed}, our training set involves 1,000 images from \textit{CAMO} and 3,040 images from \textit{COD10K}, and our test set integrates the rest from the four datasets.

\noindent \textbf{Metrics.} Following previous methods~\citep{fan2020camouflaged,fan2021concealed}, we employ four commonly-used metrics, including mean absolute error $(M)$, adaptive F-measure $(F_\beta)$,
mean E-measure $(E_\phi)$,
and structure measure $(S_\alpha)$.
Note that smaller $M$ or larger $F_\beta$, $E_\phi$, $S_\alpha$ signify better performance.

\renewcommand{\thefigure}{6}

\begin{figure}[t]
\begin{minipage}[c]{0.47\textwidth}
\centering
		\begin{subfigure}{0.237\textwidth}
		\centering
		\includegraphics[width=\textwidth]{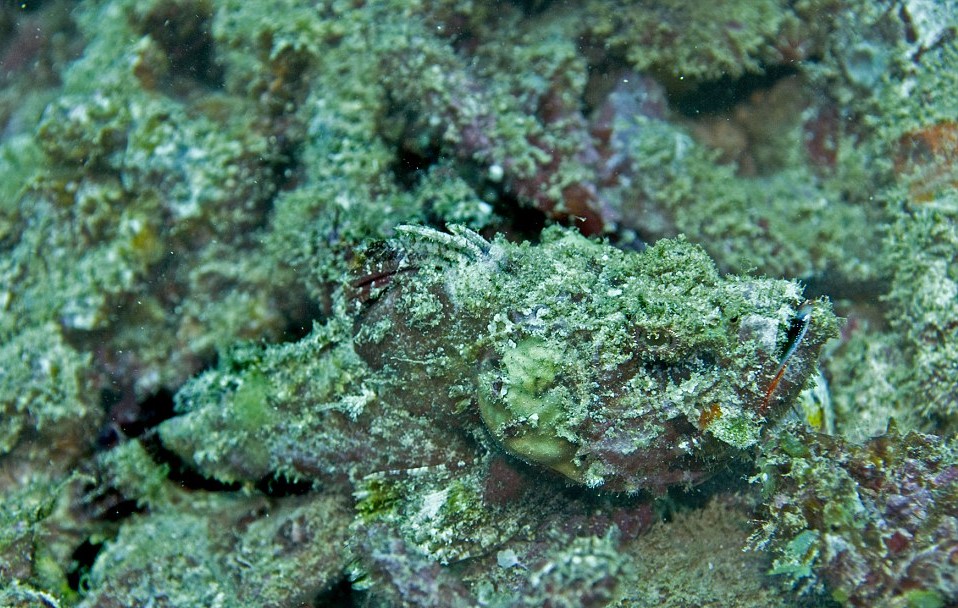}\vspace{-1pt}
	\end{subfigure}
	\hfill
	\begin{subfigure}{0.237\textwidth}  
		\centering 
		\includegraphics[width=\textwidth]{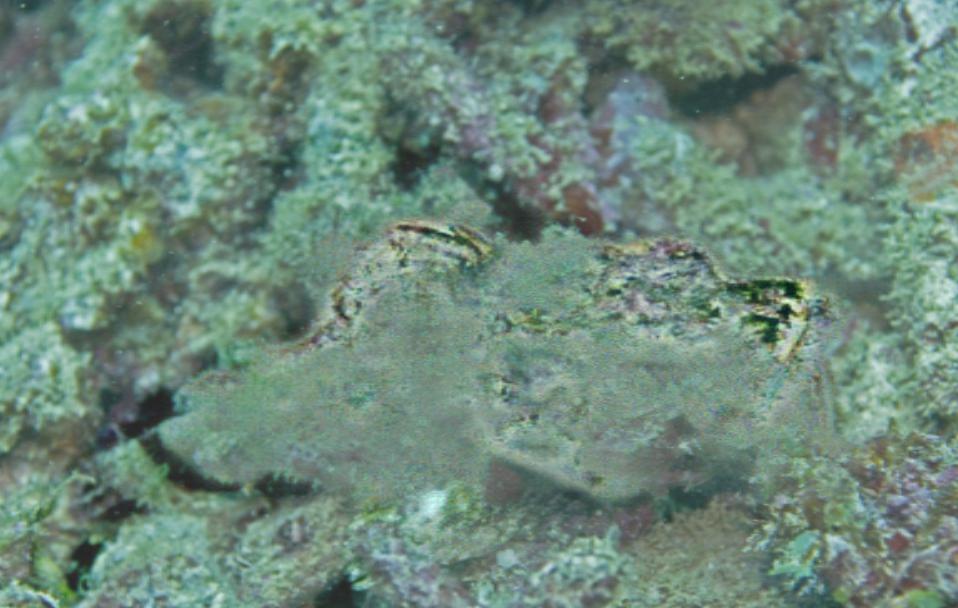}\vspace{-1pt}
	\end{subfigure}
	\hfill
	\begin{subfigure}{0.237\textwidth}  
		\centering 
		\includegraphics[width=\textwidth]{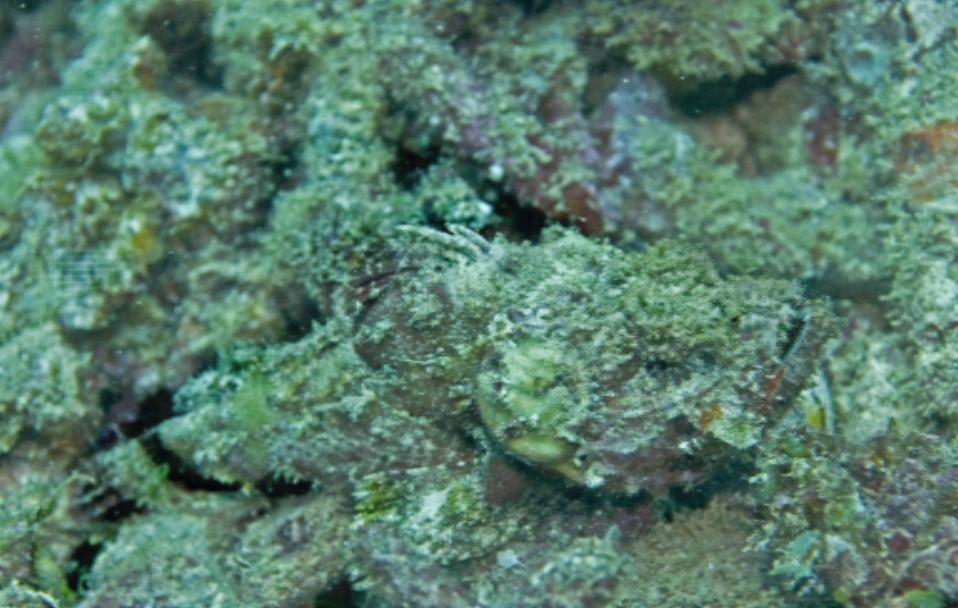}\vspace{-1pt}
	\end{subfigure}
	\hfill
	\begin{subfigure}{0.237\textwidth}   
		\centering 
		\includegraphics[width=\textwidth]{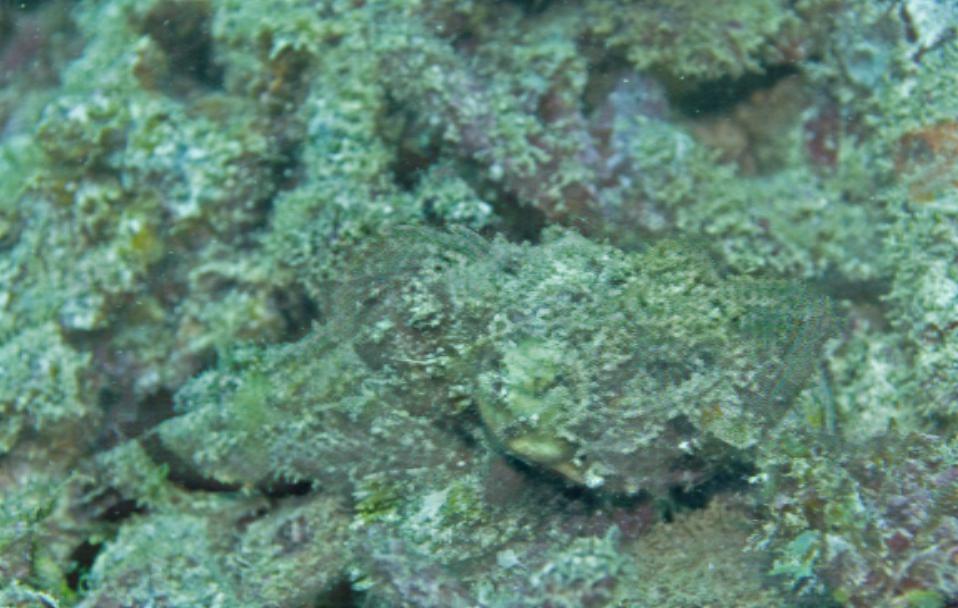}\vspace{-1pt}
	\end{subfigure} \\ \vspace{1mm}
	\begin{subfigure}{0.237\textwidth}
		\centering
		\includegraphics[width=\textwidth]{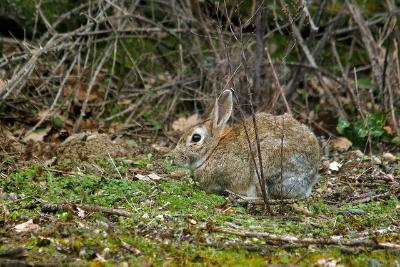}\vspace{-5pt}
		\caption{\fontsize{7.8pt}{\baselineskip}\selectfont Origin}
	\end{subfigure}
	\hfill
	\begin{subfigure}{0.237\textwidth}  
		\centering 
		\includegraphics[width=\textwidth]{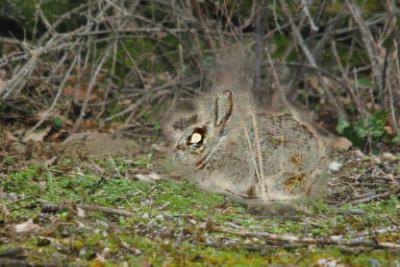}\vspace{-5pt}
		\caption{\fontsize{7.8pt}{\baselineskip}\selectfont w/o $L_g^c$}
	\end{subfigure}
	\hfill
	\begin{subfigure}{0.237\textwidth}  
		\centering 
		\includegraphics[width=\textwidth]{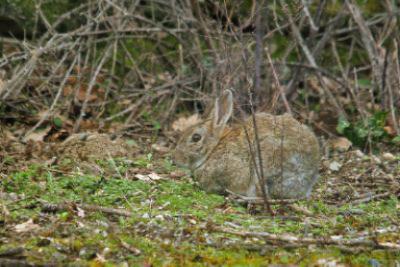}\vspace{-5pt}
		\caption{\fontsize{7.8pt}{\baselineskip}\selectfont w/ $L_f$}
	\end{subfigure}
	\hfill
	\begin{subfigure}{0.237\textwidth}   
		\centering 
		\includegraphics[width=\textwidth]{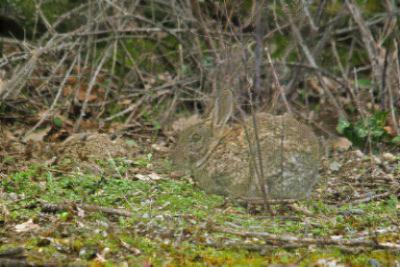}\vspace{-5pt}
		\caption{\fontsize{7.8pt}{\baselineskip}\selectfont w/ $L_g^c$} 
	\end{subfigure}\vspace{-3.5mm}
	\captionof{figure}{ \small Synthesized images of the generator trained by different losses.
 }
	\label{fig:CamouflageatorAblation}
	\vspace{-4.5mm}
\end{minipage}
\begin{minipage}[c]{0.52\textwidth}
\setlength{\abovecaptionskip}{0cm} 
	\setlength{\belowcaptionskip}{0cm}
		\centering
		\resizebox{1\columnwidth}{!}{
			\setlength{\tabcolsep}{1mm}
            \begin{tabular}{l|ccc|ccc}
            \toprule[1.5pt]
\multicolumn{1}{c|}{}& \multicolumn{3}{c|}{{Effect of $L_f$ and $L_{cl}$}} & \multicolumn{3}{c}{Effect of AT strategy} \\  \cline{2-7}
\multicolumn{1}{l|}{\multirow{-2}{*}{Metrics}} &\cellcolor{c3!20} w/o $L_g^c$ &\cellcolor{c3!20} w/ $L_f$ &\cellcolor{c3!20} w/ $L_g^c$ &\cellcolor{c3!20}  BO w/o $L_{cl}$ &\cellcolor{c3!20}  BO w/ $L_g^c$ &\cellcolor{c3!20}  AT w/ $L_g^c$ \\ \midrule
$M$~$\downarrow$ & 0.032                         & 0.030    & 0.028  & 0.030           & 0.032         & 0.028    \\
$F_\beta$~$\uparrow$ & 0.721                         & 0.750    & 0.763 & 0.752           & 0.722         & 0.763       \\
$E_\phi$~$\uparrow$ & 0.899                         & 0.907    & 0.920  & 0.910           & 0.895         & 0.920     \\
$S_\alpha$~$\uparrow$ & 0.816                         & 0.834    & 0.843 & 0.832           & 0.812         & 0.843   \\ \bottomrule[1.5pt]  
\end{tabular}}\vspace{0.5mm}		
	\captionof{table}{\small Ablation study of Camouflageator on \textit{COD10K}. $L_g^c=L_f+\lambda L_{cl}$. BO and AT are bi-level optimization and adversarial training.
	}\label{table:Camouflageator} 
	\vspace{-3mm}
\end{minipage}
\end{figure}

\subsection{Comparison with the state-of-the-arts}
\noindent\textbf{Quantitative analysis.} We compare our ICEG with 13 state-of-the-art (SOTA) solutions in three different settings. Apart from the common setting,
two other settings (multiple input scales and multiple stages) are also included, where ICEG follows the corresponding practices of ZoomNet~\citep{pang2022zoom} and SegMaR~\citep{jia2022segment}. 
As shown in~\cref{table:Qualitative}, ICEG outperforms the SOTAs by a large margin in all settings and backbones. In the common setting, ICEG overall surpasses the second-best methods in $2.1\%$, $5.2\%$, $8.3\%$ with the backbone of ResNet50 (FEDER~\citep{He2023Camouflaged}), Res2Net50 (BGNet~\citep{sun2022boundary}), Swin Transformer (ICON~\citep{zhuge2022salient}). Moreover, we also present the results of detectors optimized under Camouflageator. In~\cref{table:Qualitative}, Camouflageator generally improves other detectors by $2.8\%$ (PreyNet), $2.2\%$ (FGANet), $2.3\%$ (FEDER), and increases our ICEG by $2.5\%$ (ResNet50), $2.3\%$ (Res2Net50), $1.4\%$ (Swin Transformer), which verifies that our Camouflageator is a plug-and-play framework. Results of the compared methods are generated by their provided models with the image size of $352\times352$ for fairness.

\begin{table}[t]
        \setlength{\abovecaptionskip}{0.1cm} 
	\setlength{\belowcaptionskip}{0cm}
		\centering
		\resizebox{1\columnwidth}{!}{
			\setlength{\tabcolsep}{0.8mm}
			\begin{tabular}{l|cccccc|cccccc|c}
   \toprule[1.5pt]
\multicolumn{1}{c|}{}& \multicolumn{6}{c|}{Ablation study of CFC component} & \multicolumn{6}{c|}{Ablation study of ESD component} &Ours\\  \cline{2-14}
\multirow{-2}{*}{Metrics} &\cellcolor{c3!20} w/o CFC &\cellcolor{c3!20} w/o IFA &\cellcolor{c3!20} w/o CFA &\cellcolor{c3!20} w/o FA &\cellcolor{c3!20} w/o $L_{cc}$ &\cellcolor{c3!20} $L_{cc}$--\textgreater{}$L_{cc}^1$ & \cellcolor{c3!20}w/o ESD & \cellcolor{c3!20}w/o ESC & \cellcolor{c3!20}SC--\textgreater{}FC & \cellcolor{c3!20}SC--\textgreater{}BC & \cellcolor{c3!20}w/o AN & \cellcolor{c3!20}w/o ER & \cellcolor{c3!20}ICEG   \\ \midrule
$M$~$\downarrow$ & 0.035   & 0.032   & 0.031   & 0.033  & 0.032        & 0.032 & 0.035   & 0.034   & 0.032               & 0.031               & 0.033  & 0.034  & \textbf{0.030} \\
$F_\beta$~$\uparrow$ & 0.685   & 0.728   & 0.731   & 0.720  & 0.722        & 0.704 & 0.678   & 0.688   & 0.737               & 0.741               & 0.715  & 0.693  & \textbf{0.747} \\
$E_\phi$~$\uparrow$ & 0.866   & 0.885   & 0.893   & 0.883  & 0.887        & 0.890 & 0.864   & 0.871   & 0.896               & 0.902               & 0.890  & 0.872  & \textbf{0.906} \\
$S_\alpha$~$\uparrow$ & 0.808   & 0.814   & 0.822   & 0.812  & 0.816        & 0.813 & 0.802   & 0.806   & 0.820               & 0.822               & 0.815  & 0.804  & \textbf{0.826} \\ \bottomrule[1.5pt]
\end{tabular}}
 \caption{\small Ablation study of ICEG on \textit{COD10K}.
 ``--\textgreater{}'' is substitution. (a) FA includes both IFA and CFA. $L_{cc}^1$ is the first term of $L_{cc}$ in \cref{Eq:CCLoss}. (b) SC, FC, BC are short for separated (\cref{Eq:ESC3}), foreground (\cref{Eq:ESC3-1}), background (\cref{Eq:ESC3-2}) calibration. Note that ``w/o ER'' removes edge predictions, thus including ``w/o AN''.} 
	\label{table:AblationICEG}\vspace{-5mm}
\end{table}

\noindent\textbf{Qualitative analysis.} 
\cref{fig:qualitative} shows that 
ICEG gets more complete results than existing methods, especially for large objects whose intrinsic correlations are more dispersed (the last row). This substantiates the effectiveness of the 
our CFC module that excavates the internal coherence of camouflaged objects for generating more complete prediction maps. 
Moreover, ICEG gets clearer edges for the predictions than the existing methods, thanks to our ESD module that 
decreases uncertainty fringes and eliminates unclear edges of the segmented object.
Moreover, we can see that ICEG+ obtains even better results than ICEG, further verifying the effect of our Camouflageator framework. 

\subsection{Ablation study and analysis}
\noindent\textbf{Validity of Camouflageator.} We conduct validity analyses for Camouflageator, 
including our objective function in \cref{Eq:GeneLoss} and the adversarial training manner. As shown in~\cref{fig:CamouflageatorAblation}, the generator trained without $L_g^c$ produces the images with severe artifacts, while the one trained with fidelity loss $L_f$ only synthesizes visual-appealing images but fails to hide discriminative features. In contrast, the generator trained with our $L_g^c$ generates high-quality images with more camouflaged objects, ensuring the generalizability of the detector (see~\cref{table:Camouflageator}
). We also compare Camouflageator with the bi-level optimization (BO) framework~\citep{he2023HQG} to verify the advancement of our adversarial manner. BO involves the auxiliary generator and the detector in both the training and testing phases without adversarial losses, \textit{i.e.}, Eqs.~(\ref{Eq:AdvLoss}) and (\ref{Eq:AdvLoss1}). As the concealment loss $L_{cl}$ may limit the performance in such an end-to-end manner, we also report the results optimized without $L_{cl}$, namely with only 
$L_f$ and segmentation loss (\cref{Eq:ICENetLoss}). 
\cref{table:Camouflageator} verifies the effect of our adversarial manner.

\noindent\textbf{Effect of CFC and ESD.} 
The efficacy of CFC modules is verified in~\cref{table:AblationICEG}. In~\cref{table:AblationICEG}, we examine the impact of the CFC module (in (a)) and investigate the effect of individual components in CFC, including feature aggregation components (in (b), (c), and (d)), and $L_{cc}$ (in (e) and (f)). 
As shown in~\cref{table:AblationICEG} (e), our camouflaged consistency loss $L_{cc}$ generally improves our detector by $3.4\%$, which its positive effect. Furthermore, we demonstrate the superiority of $L_{cc}$ by incorporating $L_{cc}$ into existing cutting-edge detectors, as detailed in the supplementary materials. Additionally, we present detailed ablation results for ESD in~\cref{table:AblationICEG}, where we highlight the benefits of ESD, ESC, separated calibration, adaptive normalization, and the joint strategy to integrate the ER task into the COD task. Moreover, as observed in~\cref{table:AblationICEG}, the combination of edge guidance with foreground/background splitting using attentive masks is shown to further boost detection performance. Such discovery can bring insights for the design of edge guidance modules.

\vspace{-1mm}
\section{Conclusion}\vspace{-1.5mm}
In this paper, we propose to address COD on both the prey and predator sides. On the prey side, we introduce 
a novel adversarial training strategy, Camouflageator, to enhance the generalizability of the detector by generating more camouflaged objects harder for a COD detector to detect. On the predator side, we design a novel detector, dubbed ICEG, to address the issues of incomplete segmentation and ambiguous boundaries. In specific, ICEG employs the CFC module to excavate the internal coherence of camouflaged objects and applies the ESD module for edge prominence, thus producing complete and precise detection results. Extensive experiments verify our superiority.

\textbf{Acknowledgements.} 
This work is supported by National Key R\&D Program of China (Grant No. 2020AAA0108303), Shenzhen Science and Technology Project (Grant No. JCYJ20200109143041798). 
Shenzhen Stable Supporting Program (WDZC20200820200655001). Shenzhen Key Laboratory of next generation interactive media innovative technology (Grant No. ZDSYS 20210623092001004).
The authors express their appreciation to Dr. Fengyang Xiao for her insightful comments, improving the quality of this paper.

\clearpage

\clearpage

\bibliography{iclr2024_conference}
\bibliographystyle{iclr2024_conference}

\end{document}

%% file: math_commands.tex
\usepackage{amsmath,amsfonts,bm}

\def\eqref#1{equation~\ref{#1}}

\def\1{\bm{1}}

\DeclareMathAlphabet{\mathsfit}{\encodingdefault}{\sfdefault}{m}{sl}
\SetMathAlphabet{\mathsfit}{bold}{\encodingdefault}{\sfdefault}{bx}{n}

